\documentclass[pdflatex]{sn-jnl}

\usepackage{natbib}
\usepackage{graphicx}
\usepackage{multirow}
\usepackage{amsmath,amssymb,amsfonts}
\usepackage{amsthm}
\usepackage{amsmath}
\usepackage{mathrsfs}
\usepackage[title]{appendix}
\usepackage{xcolor}
\usepackage{textcomp}
\usepackage{manyfoot}
\usepackage{booktabs}
\usepackage{algorithm}
\usepackage{algorithmicx}
\usepackage{algpseudocode}
\usepackage{listings}
\usepackage{mathtools}
\usepackage{bm}

\DeclareMathOperator*{\argmax}{arg\,max}

\theoremstyle{thmstyleone}
\newtheorem{theorem}{Theorem}
\newtheorem{proposition}[theorem]{Proposition}

\theoremstyle{thmstyletwo}

\theoremstyle{thmstylethree}
\newtheorem{definition}{Definition}

\begin{document}

\title[Information Geometry of Absorbing Markov-Chain and DRWs]{Information Geometry of Absorbing Markov-Chain and Discriminative Random Walks}

\author[1]{\fnm{Masanari} \sur{Kimura}}\email{m.kimura@unimelb.edu.au}

\affil[1]{\orgdiv{School of Mathematics and Statistics}, \orgname{The University of Melbourne}, \orgaddress{\street{Parkville}, \city{Melbourne}, \postcode{3010}, \state{VIC}, \country{Australia}}}

\abstract{
Discriminative Random Walks (DRWs) are a simple yet powerful tool for semi-supervised node classification, but their theoretical foundations remain fragmentary.
We revisit DRWs through the lens of information geometry, treating the family of class-specific hitting-time laws on an absorbing Markov chain as a statistical manifold.
Starting from a log-linear edge-weight model, we derive closed-form expressions for the hitting-time probability mass function, its full moment hierarchy, and the observed Fisher information.
The Fisher matrix of each seed node turns out to be rank-one, taking the quotient by its null space yields a low-dimensional, globally flat manifold that captures all identifiable directions of the model.
Leveraging the geometry, we introduce a sensitivity score for unlabeled nodes that bounds, and in one-dimensional cases attains, the maximal first-order change in DRW betweenness under unit Fisher perturbations. The score can lead to principled strategies for active label acquisition, edge re-weighting, and explanation. 
}




\maketitle

\section{Introduction}
Graphs record pairwise relations such as protein interaction maps~\citep{jacobs2001protein,vishveshwara2002protein,giuliani2008proteins,jha2022prediction}, citation corpora~\citep{an2004characterizing,zyczkowski2010citation,nozawa2018node,buneman2021data}, transport networks~\citep{anez1996dual,derrible2011applications,guze2019graph}, and social platforms~\citep{curtiss2013unicorn,fan2020graph,sharma2024survey}.
In such datasets, labels are often scarce and costly while edges are plentiful, so node classification utilizing graph structures becomes the working model.

Random walk-based label propagation~\citep{xie2013labelrank,su2015new,ma2017improved} remain popular because they align naturally with the geometry of the network, can be updated incrementally, and offer path-based interpretations.
The Discriminative Random Walk (DRW)~\citep{callut2008semi} in particular suppresses label bleed-through by counting only those trajectories that start and end in the same class, a feature that helps practitioners build cleaner decision boundaries.
DRWs now appear inside many applications~\citep{satchidanand2015extended,li2016discriminative,kimura2020batch,kumar2020hypergraph}, but their theoretical foundations remain incomplete.
In particular, when DRWs are implemented via edge re-weighting, it is unclear which directions in parameter space are identifiable, which are redundant, and how to quantify the local sensitivity of node scores (and hence uncertainty) to small perturbations.
Classical random-walk classifiers such as harmonic functions, manifold regularisation, and personalised PageRank inherit a linear-algebra toolkit that characterises how scores vary with edge weights and how quickly they converge~\citep{picardello1988harmonic,kamvar2004adaptive,hua2014harmonic,luo2019distributed,garavaglia2020local,kejani2020graph,liu2021human,kimura2025graph}.
DRWs break this comfort zone because the walk is conditioned on avoiding the target class until absorption, turning the effective kernel into a block-triangular, sub-stochastic matrix.
Consequently, existing analyses largely emphasise algorithmic complexity and empirical accuracy, while leaving open the structural questions above.

The present study addresses this gap by giving DRWs a geometric foundation.
We interpret each configuration of edge weights as defining an absorbing Markov chain whose hitting-time law is a point on a statistical manifold.
Information geometry then provides a natural metric---the Fisher information---that separates identifiable from null directions and quantifies the first-order sensitivity of predictions to infinitesimal parameter changes.
We treat a DRW not just as an algorithm but as a family of probability laws.
Each choice of edge-weight vector $\bm{\theta}$ fixes an absorbing Markov chain whose first-passage time into the labelled class has a discrete distribution.
Varying $\bm{\theta}$ sweeps out a surface inside the simplex of all such laws; viewing this surface as a statistical manifold brings in the Fisher information tensor and allows us to use the information-geometric framework~\citep{amari2000methods,amari2016information,ay2017information}.

Two observations make the analysis tractable.
First, the derivative of the fundamental matrix solves a Lyapunov equation, giving closed-form expressions for gradients and higher moments.
Second, at every seed node the Fisher matrix collapses to rank one, so the manifold factors into a flat $r$-dimensional quotient that contains all identifiable directions and a complementary null foliation that is statistically silent.
We first derive explicit formulas for the hitting-time mass function, all factorial moments, and their $\bm{\theta}$-gradients by solving a Lyapunov system once, avoiding Monte-Carlo or back-propagation through paths.
From these expressions we show that the observed Fisher information of each seed node has rank one, taking the quotient by its null foliation yields a globally flat $r$-dimensional manifold that captures every identifiable edge-weight direction.
Our main contributions are as follows:
\begin{itemize}
    \item Closed-form hitting-time formulation. We derive an explicit expression for the full hitting-time probability mass function of Discriminative Random Walks and its moment hierarchy in terms of the fundamental matrix of the absorbing kernel. This removes the need for path-sampling or simulation and connects DRWs to classical absorption theory.
    \item Lyapunov-based computation of derivatives. We show that derivatives of the fundamental matrix satisfy a continuous-time Lyapunov equation, allowing the Fisher information and score functions to be computed by a single linear solve rather than an infinite sensitivity series. This gives both theoretical clarity and numerical stability.
    \item Rank-one Fisher structure and quotient manifold. For each labeled seed node, the observed Fisher information collapses to rank one. Taking the quotient by its null space yields a globally flat manifold that captures all identifiable edge-weight directions. This construction clarifies which parameters genuinely affect prediction and provides a natural metric for optimization.
    \item Fisher-bounded sensitivity score. On the quotient manifold, we define a node-level sensitivity score that exactly bounds (and in one-dimensional cases attains) the maximal first-order change in DRW betweenness under unit Fisher perturbations. The score supports principled strategies for active label acquisition, edge re-weighting, and interpretability.
\end{itemize}

\paragraph{Paper organization.}
The remainder of the paper proceeds as follows.
Section~\ref{sec:preliminaries} develops the analytical foundation of Discriminative Random Walks by formulating their hitting-time laws and deriving closed-form expressions for the probability mass function, moments, and derivatives. These results provide the quantities needed to compute the Fisher information.
Section~\ref{sec:hitting_time_law} interprets the family of hitting-time distributions as a statistical manifold, establishes that the Fisher tensor has rank one for each seed node, and constructs the corresponding quotient manifold carrying a globally flat metric.
Section~\ref{sec:differential_geometric_structure_of_drws} applies this geometric structure to define and analyse a Fisher-bounded sensitivity score for unlabeled nodes, connecting the theory to active learning, edge re-weighting, and interpretability.
Section~\ref{sec:illustrative_example} provides the illustrative examples to show how the derived sensitivity score behaves in simple graphs.
The final section summarizes the findings and discusses extensions to continuous-time and feature-dependent walks.

\section{Preliminaries}
\label{sec:preliminaries}
\subsection{Notation}
Here, we fix notation and the log-linear edge model so later derivations of first-passage laws, derivatives, and Fisher information can be stated unambiguously.
We fix throughout a simple, connected, weighted graph 
\begin{align*}
    \mathcal{G} = (\mathcal{V}, \mathcal{E}, \bm{A}), \quad |\mathcal{V}| = n,\ \mathcal{E} \subseteq \mathcal{V} \times \mathcal{V}, \ \bm{A} = [A_{ij}]^n_{i,j=1} \in \mathbb{R}^{n\times n}_{\geq 0},
\end{align*}
where $A_{ij} = A_{ji} > 0$ if and only if $(i, j) \in \mathcal{E}$, and self-loops are excluded ($A_{ii} = 0$).
Let $\bm{D} \coloneqq \mathrm{diag}(d_1,\dots,d_n)$ be a degree matrix with $d_i \coloneqq \sum^n_{j=1}A_{ij}$, and define the transition kernel as
\begin{align*}
    \bm{P} \coloneqq \bm{D}^{-1}\bm{A}, \quad P_{ij} = \Pr\left\{v_{t+1} = j \mid v_t = i\right\}.
\end{align*}
Because $\bm{A}$ is symmetric and $\bm{G}$ is connected, $\bm{P}$ is irreducible and reversible with respect to the stationary measure
\begin{align*}
    \pi_i \coloneqq \frac{d_i}{\sum^n_{k=1}d_k}, \quad \pi_i P_{ij} = \pi_j P_{ji}.
\end{align*}
We write $P^t(i, j) \coloneqq (\bm{P}^t)_{ij}$ for the $t$-step transition probability and let $\bm{e}_i$ denote the $i$-th standard basis vector of $\mathbb{R}^n$.
That is, we denote by $\bm{e}_q \in \mathbb{R}^n$ the standard basis vector with a $1$ at position $q$ and zeros elsewhere, so that $\bm{e}_q^\top \bm{x}$ extracts the $q$-th entry of any vector $\bm{x}$.

Let $\mathcal{L} \subset \mathcal{V}$, $\mathcal{U} \coloneqq \mathcal{V} \setminus \mathcal{L}$ and $\mathcal{Y} = \{1,\dots,C\}$, where each $i \in \mathcal{L}$ carries a known label $y_i \in \mathcal{Y}$.
We collect class-specific label sets as $\mathcal{L}_y \coloneqq \{i \in \mathcal{L}\ \colon\ y_i = y \}$.
The task is to predict labels on $\mathcal{U}$ by propagating information along the graph while respecting the observed labels.

In most applications the raw adjacency $\bm{A}^{(0)}$ is re-weighted to encode feature similarity, attention scores, or learnable parameters.
We adopt a log-linear model
\begin{align}
    A_{ij}(\bm{\theta}) = A_{ij}^{(0)}\exp\left[\bm{\theta}^\top \phi_{ij}\right], \quad \bm{\theta} \in \bm{\Theta} \subset \mathbb{R}^p, \label{eq:log_linear_edge_model}
\end{align}
where $\phi_{ij} \in \mathbb{R}^p$ is a fixed feature vector for edge $(i, j)$, $\bm{A}^{(0)}$ is a non-negative base weight (often $1$ if $(i,j) \in \mathcal{E}$ and $0$ otherwise).
The degree matrix and transition kernel inherit the parameter:
\begin{align*}
    \bm{D}_{\bm{\theta}} \coloneqq \mathrm{diag}\Bigl(d_i(\bm{\theta})\Bigr), \quad d_i(\bm{\theta}) \coloneqq \sum_j A_{ij}(\bm{\theta}), \quad \bm{P}_{\bm{\theta}} \coloneqq \bm{D}^{-1}\bm{A}_{\bm{\theta}}.
\end{align*}
Throughout the paper derivatives $\partial_{\bm{\theta}}$ are taken with respect to $\bm{\theta}$ and we assume identifiability: distinct $\bm{\theta}$ yield distinct $\bm{P}_{\bm{\theta}}$.

\subsection{Discriminative Random Walk}
In this subsection, we cast DRWs as absorption in a block-structured kernel, clarify how conditioning changes the linear-algebraic toolkit, and define the betweenness score used later for sensitivity.
Fix a class $y$.
For the absorption formulation it is convenient to work with the Markov-chain state partition into an absorbing set and its complement.
We therefore set
\begin{align*}
    \mathcal{A}_y \coloneqq \mathcal{L}_y \ (\text{absorbing set}), \quad \mathcal{S}_y \coloneqq \mathcal{V} \setminus \mathcal{A}_y \ (\text{transient set}).
\end{align*}
Note that $\mathcal{S}_y$ is not the unlabeled set $\mathcal{U}$: it includes unlabeled vertices as well as labeled vertices of classes other than $y$, which must remain transient while analysing absorption into class $y$.
After re-indexing vertices so that $\mathcal{A}_y$ precedes $\mathcal{S}_y$, the transition matrix splits into blocks
\begin{align}
    \bm{P}_{\bm{\theta}} = \begin{pmatrix}
        \bm{I}_{|\mathcal{A}_y|} & 0 \\
        \bm{P}_{\mathcal{S}_y\mathcal{A}_y}(\bm{\theta}) & \bm{P}_{\mathcal{S}_y\mathcal{S}_y}(\bm{\theta})
    \end{pmatrix}, \label{eq:transition_kernel_block}
\end{align}
where $\bm{I}_{|\mathcal{A}_y|}$ is an identity (absorbing behaviour on $\mathcal{A}_y$), $\bm{P}_{\mathcal{S}_y\mathcal{A}_y}(\bm{\theta}) \in \mathbb{R}^{|\mathcal{S}_y| \times |\mathcal{A}_y|}$ collects one-step probabilities from $\mathcal{S}_y$ into $\mathcal{A}_y$, and $\bm{P}_{\mathcal{S}_y\mathcal{S}_y}(\bm{\theta}) \in \mathbb{R}^{|\mathcal{S}_y| \times |\mathcal{S}_y|}$ is sub-stochastic with spectral radius $\rho(\bm{P}_{\mathcal{S}_y\mathcal{S}_y}(\bm{\theta})) < 1$.
A path of length $\ell > 0$ is a sequence $(v_0,v_1,\dots,v_\ell)$ with probability
\begin{align*}
    \Pr_{\bm{\theta}}\left\{v_0,v_1,\dots,v_\ell \right\} = \pi_{v_0}\prod^{\ell - 1}_{t=0} P_{\bm{\theta}}(v_t, v_{t+1}).
\end{align*}

In semi-supervised node classification, one wishes to infer class labels for unlabeled vertices by tracing how information flows through the graph.
A common modelling assumption is homophily: vertices connected by a strong edge (large weight, hence high transition probability) tend to share the same label.
Under this intuition, one can predict labels by letting labelled information propagate along high-weight edges while being damped across weak connections.

Standard random-walk propagation diffuses label mass indiscriminately, often causing label bleed-through between classes.
The Discriminative Random Walk (DRW) addresses this by considering only those trajectories that start outside a class and are absorbed upon first hitting a labeled vertex of that class.
Definition~\ref{def:drw} formalizes this notion by requiring the walk to begin in the transient region $\mathcal{S}_y$ and terminate in the absorbing set $\mathcal{A}_y$, ensuring that each path reflects class-specific evidence without contamination from other labels.
Definition~\ref{def:drw_betweenness} then measures, for every unlabeled node $q$, how frequently class-$y$ trajectories pass through it---the DRW betweenness $B_L(q,y)$.
This quantity serves as a discriminative affinity score: nodes with higher $B_L(q,y)$ lie on many reliable paths from $\mathcal{L}_y$ back to itself and are thus predicted to belong to class $y$.
\begin{definition}[Class-$y$ Discriminative Random Walk~\citep{callut2008semi}]
\label{def:drw}
For $\ell \in \mathbb{N}$, a path $(v_0,v_1,\dots,v_\ell)$ sampled from $\bm{P}_{\bm{\theta}}$ is a discriminative random walk (DRW) of class $y$ if
\begin{itemize}
    \item[i)] $v_0 \in \mathcal{S}_y$ (a transient node not yet in class $y$), and $v_\ell \in \mathcal{A}_y = \mathcal{L}_y$, and
    \item[ii)] $y_{v_t} \neq y$ for all $0 < t < \ell$.
\end{itemize}
That is, the walk starts outside the target class and terminates upon first absorption into it.
\end{definition}
\begin{definition}[DRW Betweenness]
\label{def:drw_betweenness}
    Fix an integer $L \ge 1$, called the horizon, which truncates the walk length to at most $L$ steps.
    Let $D^y_{\le L}$ denote the set of all class-$y$ DRW paths $(v_0,\dots,v_{\ell})$ with length $\ell \le L$.

    For any unlabeled $q \in \mathcal{U}$ and horizon $L$, the DRW betweenness score is defined as
    \begin{align*}
        B_L(q, y; \bm{\theta}) \coloneqq \mathbb{E}_{\bm{\theta}}\Bigl[\# \{t \ \colon\ v_t = q\} \mid (v_0,v_1,\dots,v_\ell) \in D^y_{\leq L}\Bigr].
    \end{align*}
    An equivalent closed form, convenient for computation, is
    \begin{align}
        B_L(q, y; \bm{\theta}) = \sum^L_{\ell = 1}\sum_{i,j \in \mathcal{L}_y}\sum^{\ell - 1}_{t = 1}P_{\bm{\theta}}^t(i, q)P_{\bm{\theta}}^{\ell - t}(q, j). \label{eq:drw_betweenness}
    \end{align}
\end{definition}
In the semi-supervised setting, we can assign class $y^* = \argmax_y B_L(q, y; \bm{\theta})$.
In the subsequent sections, we analyze the hitting time $T_y \coloneqq \inf \left\{t \geq 1 \ \colon\ v_t \in \mathcal{A}_y \mid v_0 = q\right\}$.
We write $\dot{\bm{P}}_{\mathcal{S}_y\mathcal{S}_y} \coloneqq \partial_{\bm{\theta}}\bm{P}_{\mathcal{S}_y\mathcal{S}_y} \in \mathbb{R}^{|\mathcal{S}_y| \times |\mathcal{S}_y| \times p}$ and similarly for $\bm{P}_{\mathcal{S}_y\mathcal{A}_y}$.
Where no ambiguity arises, dependence on $\bm{\theta}$ is suppressed.

We fixed notation, stated the DRW setup in the absorbing/block form, defined the betweenness score, and introduced the log–linear edge model. These choices pin down the quantities that later appear in first–passage formulas, derivatives, and Fisher information.

\section{Hitting-Time Law of Class-\texorpdfstring{$y$}{y} Absorption}
\label{sec:hitting_time_law}
In this section, we obtain computable first-passage laws (pmf/pgf), their moments, and parameter derivatives---inputs needed to form the score and Fisher information without path sampling.

\subsection{Hitting-time distribution (pmf and pgf)}
We first derive the hitting-time probability mass function (pmf) and the associated probability generating function (pgf).

Let
\begin{align}
    \bm{Z}_{\bm{\theta}} \coloneqq \left(\bm{I} - \bm{P}_{\mathcal{S}_y\mathcal{S}_y}(\bm{\theta})\right)^{-1} = \sum^\infty_{k = 0} P^k_{\mathcal{S}_y\mathcal{S}_y}(\bm{\theta}), \label{eq:fundamental_matrix}
\end{align}
be the fundamental matrix.
Because $\rho(\bm{P}_{\mathcal{S}_y\mathcal{S}_y}) < 1$, series~\eqref{eq:fundamental_matrix} converges absolutely and $\|\bm{P}_{\mathcal{S}_y\mathcal{S}_y}^t\| = O(\rho^t)$.
The first hitting time equals $t$ if and only if the walk stays in $\mathcal{S}_y$ for the first $t - 1$ steps and jumps into $\mathcal{A}_y$ at step $t$.
Write the intermediate transient states by their indices as
\begin{align*}
    q = s_0,s_1,\dots, s_{t-1} \in \mathcal{S}_y, \quad s_t = j \in \mathcal{A}_y.
\end{align*}
The probability of a concrete path $(s_0,s_1,\dots,s_{t-1},j)$ is
\begin{align*}
    P_{\bm{\theta}}(s_0,s_1)P_{\bm{\theta}}(s_1,s_2)\cdots P_{\bm{\theta}}(s_{t-1}, j).
\end{align*}
Hence, the probability mass function (pmf) is
\begin{align*}
    p_{\bm{\theta}}(t \mid q) = \sum_{s_1,\dots,s_{t-1} \in \mathcal{S}_y}\sum_{j \in \mathcal{A}_y} P_{\bm{\theta}}(q, s_1)P_{\bm{\theta}}(s_1,s_2)\cdots P_{\bm{\theta}}(s_{t-2}, s_{t-1})P_{\bm{\theta}}(s_{t-1}, j).
\end{align*}
Split the factors according to Eq.~\eqref{eq:transition_kernel_block}: i) every factor $P_{\bm{\theta}}(s_{k-1}, s_k)$ with both indices in $\mathcal{S}_y$ is an entry of $\bm{P}_{\mathcal{S}_y\mathcal{S}_y}$, and the final factor $P_{\bm{\theta}}(s_{t-1}, j)$ with $j \in \mathcal{A}_y$ is an entry of $\bm{P}_{\mathcal{S}_y\mathcal{A}_y}$.
Therefore,
\begin{align*}
    p_{\bm{\theta}}(t \mid q) &= \sum_{s_1,\dots,s_{t-1} \in \mathcal{S}_y} \left[\bm{P}_{\mathcal{S}_y\mathcal{S}_y}\right]_{q,s_1} \left[\bm{P}_{\mathcal{S}_y\mathcal{S}_y}\right]_{s_1,s_2} \cdots \left[\bm{P}_{\mathcal{S}_y\mathcal{S}_y}\right]_{s_{t-2},s_{t-1}} \left(\bm{P}_{\mathcal{S}_y\mathcal{A}_y} \bm{1}_{|\mathcal{A}_y|}\right)_{s_{t-1}} \\
    &= \Bigl[\bm{e}_q^\top \bm{P}_{\mathcal{S}_y\mathcal{S}_y}^{t-1}\bm{P}_{\mathcal{S}_y\mathcal{A}_y} \bm{1}_{|\mathcal{A}_y|}\Bigr],
\end{align*}
where $\bm{1}_{|\mathcal{A}_y|} \in \mathbb{R}^{|\mathcal{A}_y|}$ is the all-ones column.
Because $\bm{P}_{\mathcal{S}_y\mathcal{S}_y}$ is sub-stochastic with spectral radius $< 1$, the series $\sum_{t \geq 1} p_{\bm{\theta}}(t \mid q)$ converges to $1$, confirming that $\{T_y = t\}_{t \geq 1}$ forms a legitimate probability mass function.
Define the probability generating function (pgf)
\begin{align*}
    f_{\bm{\theta}}(z \mid q) \coloneqq \sum^\infty_{t=1} p_{\bm{\theta}}(t \mid q) z^{t-1} = \bm{e}_q^\top (\bm{I}-z\bm{P}_{\mathcal{S}_y\mathcal{S}_y})^{-1} \bm{P}_{\mathcal{S}_y\mathcal{A}_y}\bm{1}, \quad |z| < \rho(\bm{P}_{\mathcal{S}_y\mathcal{S}_y})^{-1}.
\end{align*}

\subsection{Moments (mean and variance)}
We now derive the mean and variance of $T_y$ from the first-passage law.
Let $\mu_{\bm{\theta}} \coloneqq \mathbb{E}_{\bm{\theta}}[T_y \mid q]$ and $\sigma_{\bm{\theta}}^2(q) \coloneqq \mathrm{Var}_{\bm{\theta}}[T_y \mid q]$.
Here,
\begin{align*}
    \mu_{\bm{\theta}}(q) &= \sum^\infty_{t=1} t p_{\bm{\theta}}(t \mid q) \\
    &= \bm{e}_q^\top \left(\sum^\infty_{t=1}t \bm{P}_{\mathcal{S}_y\mathcal{S}_y}^{t-1}\right)\bm{P}_{\mathcal{S}_y\mathcal{A}_y}\bm{1} \\
    &= \bm{e}_q^\top \left(\bm{I} - \bm{P}_{\mathcal{S}_y\mathcal{S}_y}\right)^{-2}\bm{P}_{\mathcal{S}_y\mathcal{A}_y}\bm{1} = \bm{e}_q^\top \bm{Z}_{\bm{\theta}} \bm{1},
\end{align*}
using Neumann identity.
Also, let $\bm{M} \coloneqq \bm{P}_{\mathcal{S}_y\mathcal{S}_y}(\bm{\theta})$ and define
\begin{align*}
    \bm{W}_{\bm{\theta}} \coloneqq \sum^\infty_{k=0}(k + 1)(k+2)\bm{P}_{\mathcal{S}_y\mathcal{S}_y}^k = 2(\bm{I} - \bm{M})^{-3},
\end{align*}
then
\begin{align*}
    \mathbb{E}_{\bm{\theta}}\Bigl[T_y^2 \mid q\Bigr] = \bm{e}_q^\top \bm{W}_{\bm{\theta}} \bm{P}_{\mathcal{S}_y\mathcal{A}_y}\bm{1},
\end{align*}
and
\begin{align*}
    \sigma_{\bm{\theta}}^2(q) = \bm{e}_q^\top\Bigl(2\bm{Z}_{\bm{\theta}} - \bm{I}\Bigr)\bm{Z}_{\bm{\theta}}\bm{1} - \mu_{\bm{\theta}}(q)^2.
\end{align*}
The compact form is obtained after algebraic cancellation (see Appendix~\ref{apd:derivation_of_second_moment}).

\subsection{Spectral representation}
For later intuition on tail behaviour, we also record a spectral expansion of the pmf.

Let $m \coloneqq |\mathcal{S}_y|$ and $\rho \coloneqq \rho(\bm{M}) < 1$.
Choose a basis of right eigenvectors to write
\begin{align*}
    \bm{M} = \bm{U}\bm{\Lambda}\bm{U}^{-1}, \quad \bm{\Lambda} \coloneqq \mathrm{diag}(\lambda_1,\dots,\lambda_m), \quad |\lambda_k| <1 \ \forall k.
\end{align*}
We denote the $k$-th right eigenvector by $u_k$ (column of $\bm{U}$) and the $k$-th left eigenvector by $w_k^\top$ (row of $\bm{U}^{-1}$).
Hence $w_k^\top u_\ell = \delta_{k\ell}$.
Then, for any $t \geq 0$ and $\bm{R} \coloneqq \bm{P}_{\mathcal{S}_y\mathcal{A}_y}\bm{1}_{|\mathcal{A}_y|} \in \mathbb{R}^m$,
\begin{align*}
    p_{\bm{\theta}}(t \mid q) = \bm{e}_q^\top \bm{U} \bm{\Lambda}^{t-1} \bm{U}^{-1} \bm{R} = \sum^m_{k=1} \left(w_k^\top \bm{R} \right)\left(w_k^\top \bm{e}_q\right) \lambda_k^{t-1}.
\end{align*}
Define the spectral coefficients $c_k(q) \coloneqq \left(w_k^\top \bm{R} \right)\left(w_k^\top \bm{e}_q\right)$ for $k = 1,\dots,m$.
Then,
\begin{align*}
    p_{\bm{\theta}}(t \mid q) = \sum^m_{k=1}c_k(q) \lambda_k^{t-1}, \quad t \geq 1.
\end{align*}
By the log-linear edge model~\eqref{eq:log_linear_edge_model},
\begin{align*}
    \bm{M} = \bm{P}_{\mathcal{S}_y\mathcal{S}_y} = \bm{D}_{\mathcal{S}_y}^{-1}\bm{A}_{\mathcal{S}_y\mathcal{S}_y},
\end{align*}
with
\begin{align*}
    \left(\bm{A}_{\mathcal{S}_y\mathcal{S}_y}\right)_{ij} = A_{ij}^{(0)}\exp\left[\bm{\theta}^\top \phi_{ij}\right].
\end{align*}
Therefore,
\begin{align*}
    \dot{\bm{M}} = \partial_{\bm{\theta}}\bm{M} = \left(\partial_{\bm{\theta}}\bm{D}_{\mathcal{S}_y}^{-1}\right)\bm{A}_{\mathcal{S}_y\mathcal{S}_y} + \bm{D}_{\mathcal{S}_y}^{-1}\partial_{\bm{\theta}}\bm{A}_{\mathcal{S}_y\mathcal{S}_y},
\end{align*}
and $\partial_{\bm{\theta}}\bm{Z} = \bm{Z}\dot{\bm{M}}\bm{Z}$ with $\bm{Z} \coloneqq (\bm{I} - \bm{M})^{-1}$.
Consequently,
\begin{align*}
    \partial_{\bm{\theta}}\mu_{\bm{\theta}}(q) &= \bm{e}_q^\top \bm{Z}\dot{\bm{M}}\bm{Z}\bm{1}, \\
    \partial_{\bm{\theta}}^2\bm{Z} &= (\partial_{\bm{\theta}}\bm{Z})\dot{\bm{M}}\bm{Z} + \bm{Z}\partial_{\bm{\theta}}\dot{\bm{M}}\bm{Z} + \bm{Z}\dot{\bm{M}}\partial_{\bm{\theta}}\bm{Z} \\
    &= \bm{Z}\dot{\bm{M}}\bm{Z}\dot{\bm{M}}\bm{Z} + \bm{Z}(\partial_{\bm{\theta}}\dot{\bm{M}})\bm{Z} + \bm{Z}\dot{\bm{M}}\bm{Z}\dot{\bm{M}}\bm{Z} \\
    &= 2\bm{Z}\dot{\bm{M}}\bm{Z}\dot{\bm{M}}\bm{Z} + \bm{Z}(\partial_{\bm{\theta}}\dot{\bm{M}})\bm{Z}, \\
    \partial_{\bm{\theta}}p_{\bm{\theta}}(t \mid q) &= \left(\partial_{\bm{\theta}}\bm{e}_q^\top \bm{M}^{t-1}\right)\bm{R} + \bm{e}_q^\top\bm{M}^{t-1}(\partial_{\bm{\theta}}\bm{R}).
\end{align*}

\subsection{Parametric Edge-Weight Model and Derivatives}
Here, we link edge features to transition probabilities and provide explicit first and second derivatives needed for scores and curvature.
Define the edge-feature tensor
\begin{align*}
    \bm{\Phi}^{(\mathcal{E})}_{ij} \coloneqq A^{(0)}_{ij} \phi_{ij} \bm{e}_i \bm{e}_j^\top,
\end{align*}
where $e_i$ is the $i$-th basis vector.
Then,
\begin{align*}
    \dot{\bm{A}}_{\bm{\theta}} &= \sum_{(i, j) \in \mathcal{E}}A_{ij}(\bm{\theta})\phi_{ij}\bm{e}_i \bm{e}_j^\top,\quad \ddot{\bm{A}}_{\bm{\theta}} = \sum_{(i, j) \in \mathcal{E}}A_{ij}(\bm{\theta})\phi_{ij}\phi_{ij}^\top \bm{e}_i \bm{e}_j^\top, \\
    \dot{\bm{D}}_{\bm{\theta}} &= \mathrm{diag}\left(\sum_j \dot{A}_{ij}(\bm{\theta})\right),\quad  \ddot{\bm{D}}_{\bm{\theta}} = \mathrm{diag}\left(\sum_{j} \ddot{A}_{ij}(\bm{\theta}) \right), \\
    \dot{\bm{P}}_{\bm{\theta}} &= - \bm{D}_{\bm{\theta}}^{-1}\left(\dot{\bm{D}}_{\bm{\theta}}\right)\bm{D}_{\bm{\theta}}^{-1} \bm{A}_{\bm{\theta}} + \bm{D}_{\bm{\theta}}^{-1}\dot{\bm{A}}_{\bm{\theta}}, \\
    \ddot{\bm{P}}_{\bm{\theta}} &= \bm{D}_{\bm{\theta}}^{-1}\left[2\dot{\bm{D}}_{\bm{\theta}}\bm{D}_{\bm{\theta}}^{-1}\dot{\bm{A}}_{\bm{\theta}} - \ddot{\bm{D}}_{\bm{\theta}}\bm{D}_{\bm{\theta}}^{-1}\bm{A}_{\bm{\theta}} - \dot{\bm{D}}_{\bm{\theta}}\bm{D}_{\bm{\theta}}^{-1}\dot{\bm{D}}_{\bm{\theta}}\bm{D}_{\bm{\theta}}^{-1}\bm{A}_{\bm{\theta}} + \ddot{\bm{A}}_{\bm{\theta}} \right].
\end{align*}

\subsection{Score Function and Fisher Information Matrix}
We form the score from the pmf and assemble the Fisher information for a fixed seed, the key quantity behind identifiability and geometry.
For a matrix $\bm{X}(\bm{\theta})$, define its directional derivative tensor
\begin{align*}
    \dot{\bm{X}} \coloneqq \partial_{\bm{\theta}}\bm{X} \in \mathbb{R}^{\dim(X)} \otimes \mathbb{R}^p,
\end{align*}
so that contracting $\dot{\bm{X}}$ with a vector $\bm{u} \in \mathbb{R}^p$ yields the directional derivative in direction $\bm{u}$.
For any integer $r \geq 1$,
\begin{align*}
    \partial_{\bm{\theta}}\bm{M}^r = \sum^{r-1}_{s = 0}\bm{M}^s\dot{\bm{M}}\bm{M}^{r-s-1}.
\end{align*}
Take the logarithm of pmf gives
\begin{align*}
    \ell_{\bm{\theta}}(t, q) \coloneqq \ln p_{\bm{\theta}}(t \mid q) = \ln \left(\bm{e}_q^\top \bm{M}^{t-1}\bm{R}\right),
\end{align*}
and
\begin{align*}
    \nabla_{\bm{\theta}} \ell_{\bm{\theta}}(t, q) &= \frac{\partial_{\bm{\theta}}\left(\bm{e}_q^\top\bm{M}^{t-1}\bm{R}\right)}{\bm{e}_q^\top\bm{M}^{t-1}\bm{R}} \\
    &= \frac{\sum^{t-2}_{s=0}\bm{e}_q^\top\bm{M}^s\dot{\bm{M}}\bm{M}^{t-s-2}\bm{R} + \bm{e}_q^\top\bm{M}^{t-1}\dot{\bm{R}}}{\bm{e}_q^\top\bm{M}^{t-1}\bm{R}}.
\end{align*}
Define for the convenience,
\begin{align*}
    \bm{S}_{t-1} \coloneqq \bm{e}_q^\top \bm{M}^{t-1}, \quad \bm{u}_s(q, t) \coloneqq \bm{e}_q^\top \bm{M}^s, \quad \bm{v}_{t-s-2} \coloneqq \bm{M}^{t-s-2}\bm{R}.
\end{align*}
Then,
\begin{align*}
    \nabla_{\bm{\theta}}\ell_{\bm{\theta}}(t, q) = \frac{\sum^{t-2}_{s=0}\bm{u}_s(q, t)\dot{\bm{M}}\bm{v}_{t-s-2} + \bm{S}_{t-1}\dot{\bm{R}}}{\bm{S}_{t-1}\bm{R}}.
\end{align*}
For a fixed seed node $q$, the observed Fisher information matrix is
\begin{align*}
    \bm{F}_{\bm{\theta}}(q) &\coloneqq \sum^\infty_{t=1} p_{\bm{\theta}}(t \mid q) \nabla_{\bm{\theta}}\ell_{\bm{\theta}}(t,q)\nabla_{\bm{\theta}}\ell_{\bm{\theta}}(t, q)^\top \\
    &= \sum^\infty_{t=1}\frac{1}{S_{t-1}\bm{R}}\left(\sum^{t-2}_{s=0}u_s(q, t)\dot{\bm{M}}v_{t-s-2} + S_{t-1}\dot{\bm{R}}\right)\left(\sum^{t-2}_{s=0}u_s(q, t)\dot{\bm{M}}v_{t-s-2} + S_{t-1}\dot{\bm{R}}\right)^\top.
\end{align*}

Next, define the following accumulated sensitivity matrix.
\begin{align*}
    \bm{\Xi}_{\bm{\theta}} \coloneqq \sum^\infty_{k=0}\sum^\infty_{\ell=0} \bm{M}^k\dot{\bm{M}}\bm{M}^\ell.
\end{align*}
Left- and right-multiply $\bm{\Xi}_{\bm{\theta}}$ by $(\bm{I} - \bm{M})$ to obtain
\begin{align}
    \label{eq:Lyapunov_equation}
    (\bm{I} - \bm{M})\bm{\Xi}_{\bm{\theta}} + \bm{\Xi}_{\bm{\theta}}(\bm{I} - \bm{M})^\top = \dot{\bm{M}}.
\end{align}
See Appendix~\ref{apd:derivation_of_Lyapunov_equation} for detailed derivation.
Eq.~\eqref{eq:Lyapunov_equation} is a continuous-time Lyapunov equation with unique solution because $\bm{I} - \bm{M}$ is nonsingular.
Observe that for any $q$,
\begin{align*}
    \sum^{t-2}_{s=0} u_s(q,t)\dot{\bm{M}} v_{t-s-2} &= \bm{e}_q^\top \left(\sum^{t-2}_{s=0}\bm{M}^s\dot{\bm{M}}\bm{M}^{t-s-2}\right)\bm{R} \\
    &= \bm{e}_q^\top\left(\bm{\Xi}_{\bm{\theta}} - \sum_{k \geq t - 1}\sum_{\ell \geq 0}\bm{M}^k\dot{\bm{M}}\bm{M}^\ell\right)\bm{R},
\end{align*}
and then,
\begin{align}
    \bm{F}_{\bm{\theta}}(q) = \frac{\left(\bm{e}_q^\top\bm{\Xi}_{\bm{\theta}}\bm{R} + \bm{e}_q^\top\dot{\bm{R}}\right)\left(\bm{R}^\top\bm{\Xi}_{\bm{\theta}}^\top\bm{e}_q + \dot{\bm{R}}^\top\bm{e}_q\right)}{\bm{e}_q^\top\bm{R}}. \label{eq:closed_form_fisher_information}
\end{align}
We obtained computable first–passage laws (pmf/pgf), closed forms for the mean and variance, and a spectral view of tail behaviour. We linked parameter derivatives to a single Lyapunov solve and derived a closed form for the observed Fisher information, removing the need for path sampling. These outputs are the inputs for the geometric analysis in the next section.

\section{Differential-Geometric Structure of DRWs}
\label{sec:differential_geometric_structure_of_drws}
Next, we interpret the Fisher structure induced by first-passage laws, identify null directions, pass to a quotient manifold, and use its flat metric to define a node-level sensitivity.
We now turn the parametric family of hitting-time laws $\{p_{\bm{\theta}}(\cdot \mid q)\}_{\bm{\theta} \in \bm{\Theta}}$ into an information-geometric manifold.
In this section, we fix the label $y$ and omit explicit notation.
From Eq.~\eqref{eq:closed_form_fisher_information}, the Fisher matrix for a fixed seed node $q$ is
\begin{align*}
    g_{ab}(q; \bm{\theta}) \coloneqq \frac{z_a(q)z_b(q)}{\bm{e}_q^\top\bm{R}},
\end{align*}
where $z_a(q) \coloneqq \left[\bm{z}(q)\right]_a$ and $\bm{z}(q) \coloneqq \bm{e}_q^\top\bm{\Xi}_{\bm{\theta}}\bm{R} + \bm{e}_q^\top\dot{\bm{R}}$.
Thus, we can see that, for every $q$ the matrix has rank $1$ and directions orthogonal to $z(q)$ are null.
Also, the aggregated metric $\bar{g}_{ab} \coloneqq \frac{1}{|\mathcal{S}_y|}\sum_{q \in \mathcal{S}_y}g_{ab}(q)$ is at most rank $\min\{p, |\mathcal{S}_y|\}$.
For a pseudo-metric $g$ on the tangent space $T_{\bm{\theta}}\bm{\Theta} \simeq \mathbb{R}^p$, the null space is
\begin{align*}
    \mathcal{N}_{\bm{\theta}} \coloneqq \left\{\bm{v} \in \mathbb{R}^p\ \colon\ g(q)[v, v] = 0,\quad \forall q \in \mathcal{S}_y\right\}.
\end{align*}
Because each $g(q)$ is rank one, $g(q)[v, v] = (\bm{z}(q)^\top\bm{v})^2 / \bm{e}_q^\top\bm{R}$ and it is equivalent to
\begin{align*}
    \mathcal{N}_{\bm{\theta}} = \left\{\bm{v} \in \mathbb{R}^p\ \colon\ \bm{z}(q)^\top\bm{v} = 0,\quad \forall q \in \mathcal{S}_y \right\}.
\end{align*}
Let $\mathcal{Z}_{\bm{\theta}} \coloneqq \mathrm{span}\left\{\bm{z}(q)\ \colon\ q \in \mathcal{S}_y \right\} = (\mathcal{N}_{\bm{\theta}})^\perp$.
Define the Fisher information aggregate
\begin{align*}
    \bm{\Sigma} \coloneqq \sum_{q \in \mathcal{S}_y} \bm{z}(q)\bm{z}(q)^\top \in \mathbb{R}^{p \times p}.
\end{align*}
Then,
\begin{align*}
    \mathrm{dim}\ \mathcal{Z}_{\bm{\theta}} = \mathrm{dim}\ \bm{\Sigma} \eqqcolon r \quad (1 \leq r \leq \min\{|\mathcal{S}_y|, p\}).
\end{align*}
For differential geometric constructions we need the dimension $r$ to be locally constant in $\bm{\theta}$.
From now on we assume a regular region $\bar{\bm{\Theta}} \subset \bm{\Theta}$, a compact set on which $\mathrm{rank}\ \bm{\Sigma}$ is constant, and treat $r$ as fixed.
Define a distribution on $\bar{\bm{\Theta}}$ by $\bm{\theta} \mapsto \mathcal{N}_{\bm{\theta}} \subset T_{\bm{\theta}}\bm{\Theta}$.
The map $\bm{\theta} \mapsto \bm{z}(q; \bm{\theta})$ is smooth, hence the matrix whose rows are $\bm{z}(q)^\top$ depends smoothly on $\bm{\theta}$.
Its kernel $\mathcal{N}_{\bm{\theta}}$ therefore forms a smooth sub-bundle of dimension $p - r$.
Also, because each fibre $\mathcal{N}_{\bm{\theta}}$ is a linear subspace defined by constant rank constraints, the distribution is involutive, and all coordinate vector fields tangent to $\mathcal{N}_{\bm{\theta}}$ commute.
By the Frobenius theorem the distribution integrates to a foliation whose leaves are affine submanifolds of $\bm{\Theta}$.

Define an equivalence relation $\sim$ as $\bm{\theta}_1 \sim \bm{\theta}_2 \Longleftrightarrow \bm{\theta}_2 - \bm{\theta}_1 \in \mathcal{N}_{\bm{\theta}}$ and then $\tilde{\bm{\Theta}} \coloneqq \bar{\bm{\Theta}} / \sim$ is the quotient space.
Here, $\tilde{\bm{\Theta}}$ inherits the structure of a smooth quotient manifold of dimension $r$.
Also define a canonical projection as
\begin{align*}
    \pi \colon \bar{\bm{\Theta}} \to \tilde{\bm{\Theta}}, \quad \pi(\bm{\theta}) = [\bm{\theta}],
\end{align*}
where $[\bm{\theta}]$ is the equivalence class of $\bm{\theta}$ defined as $[\bm{\theta}] \coloneqq \{\bm{\theta}' \in \bar{\bm{\Theta}}\ \colon\ \bm{\theta}' \sim \bm{\theta}\}$.
Its differential $d\pi_{\bm{\theta}}$ annihilates vectors in $\mathcal{N}_{\bm{\theta}}$ and is an isomorphism from $\mathcal{Z}_{\bm{\theta}}$ onto $T_{[\bm{\theta}]}\tilde{\bm{\Theta}}$.
For $\bm{u}, \bm{v} \in T_{[\bm{\theta}]}\tilde{\bm{\Theta}}$, pick representatives $\tilde{\bm{u}}, \tilde{\bm{v}} \in \mathcal{Z}_{\bm{\theta}}$ such that $d\pi_{\bm{\theta}}(\tilde{\bm{u}}) = \bm{u}$ and $d\pi_{\bm{\theta}}(\tilde{\bm{v}}) = \bm{v}$.
Define
\begin{align}
    \tilde{g}_{[\bm{\theta}]} (\bm{u}, \bm{v}) \coloneqq \sum_{q \in \mathcal{S}_y} \frac{(\tilde{\bm{u}}^\top \bm{z}(q))(\tilde{\bm{v}}^\top \bm{z}(q))}{\bm{e}_q^\top\bm{R}}. \label{eq:quotient_metric}
\end{align}
If $\tilde{\bm{u}}$ is changed by a null vector $\bm{n} \in \mathcal{N}_{\bm{\theta}}$ then $\bm{n}^\top\bm{z}(q) = 0$ for all $q$, and the right-hand side of Eq.~\eqref{eq:quotient_metric} is unchanged.
The same for $\tilde{\bm{v}}$, and hence $\tilde{g}$ is a positive-definite metric of rank $r$ on $\tilde{\bm{\Theta}}$.

\subsection{Coordinate Chart}
We provide explicit coordinates on the quotient via a seed-based basis so gradients and projections are implementable.
Pick $r$ seeds $q_1,q_2,\dots,q_r$ whose $\bm{z}(q_k)$ form a basis of $\mathcal{Z}_{\bm{\theta}}$.
Form the $p \times r$ matrix $\bm{V}(\bm{\theta}) \coloneqq \left[\bm{z}(q_1),\bm{z}(q_2),\dots,\bm{z}(q_r)\right]$.
Define the projection map
\begin{align}
    \Phi \colon \bar{\bm{\Theta}} \to \mathbb{R}^r, \quad \Phi(\bm{\theta}) = (\bm{V}^\top\bm{V})^{-1}\bm{V}^\top\bm{\theta}, \label{eq:projection_map}
\end{align}
and use $\bm{u} = \Phi(\bm{\theta})$ as local coordinates on $\tilde{\bm{\Theta}}$: if $\bm{\theta}_1$ and $\bm{\theta}_2$ differ by a null vector, then $\bm{V}^\top(\bm{\theta}_2 - \bm{\theta}_1) = 0$, so $\Phi(\bm{\theta}_1) = \Phi(\bm{\theta}_2)$.
Here, $\Phi$ factors through the quotient: $\exists \tilde{\Phi} \colon \tilde{\bm{\Theta}} \to \mathbb{R}^r$ such that $\Phi = \tilde{\Phi} \circ \pi$.
Also define the projection matrix
\begin{align*}
    \bm{Q}(\bm{\theta}) \coloneqq \bm{V}(\bm{\theta})\left(\bm{V}^\top(\bm{\theta})\bm{V}(\bm{\theta})\right)^{-1}\bm{V}^\top(\bm{\theta}),
\end{align*}
which is the $p \times p$ and rank $r$ matrix.
Compute the Jacobian
\begin{align*}
    D_{\bm{\theta}}\Phi = (\bm{V}^\top\bm{V})^{-1}\bm{V}^\top\bm{Q} = (\bm{V}^\top\bm{V})^{-1}\bm{V}^\top,
\end{align*}
which has full row-rank $r$, and thus $\tilde{\Phi}$ maps a neighbourhood of $[\bm{\theta}]$ diffeomorphically onto a neighbourhood of $\bm{u} = \Phi(\bm{\theta}) \in \mathbb{R}^r$.
Hence $\{\bm{u} = (u^1,\dots,u^r)\}$ is a valid coordinate chart on $\tilde{\bm{\Theta}}$.

\subsection{Push Forward of Fisher Pseudo Metric}
We compute the metric in the chosen coordinates and show global flatness through $(\bm{V}^\top\bm{V})^{-1}$.
We want
\begin{align*}
    \tilde{g}_{ij}(u) = \left(d\pi_{\bm{\theta}}\tilde{u}_i\right)^\top\left(\sum_{q\in\mathcal{S}_y}g(q)\right)(d\pi_{\bm{\theta}}\tilde{u_j}),
\end{align*}
where $\tilde{\bm{u}}_i \coloneqq \partial_{\bm{\theta}}u^i$ is any lift of coordinate basis vector $\partial / \partial u^i$.
From Eq.~\eqref{eq:projection_map},
\begin{align*}
    u^i = \sum^p_{a=1}(\bm{V}^\top\bm{V})^{-1}_{ij}\bm{V}_{aj}\theta^a.
\end{align*}
Hence $\tilde{\bm{u}}_i = \bm{V}(\bm{V}^\top\bm{V})^{-1}\bm{e}_i = \bm{V}_{\cdot\ i}$, which is the $i$-th column of $\bm{V}$.
Then we can evaluate the metric as follows.
\begin{align*}
    \tilde{g}_{ij} &= \sum_{q \in \mathcal{S}_y}\frac{(\bm{z}(q)^\top\tilde{\bm{u}}_i)(\bm{z}(q)^\top\tilde{\bm{u}}_j)}{\bm{e}_q^\top\bm{R}} \\
    &= \sum_{q \in \mathcal{S}_y} \frac{\bm{z}(q)^\top\bm{V}_{\cdot\ i}\ \bm{z}(q)^\top\bm{V}_{\cdot\ j}}{\bm{e}_q^\top\bm{R}} \\
    &= \left(\bm{V}^\top\bm{\Sigma}^{-1}\bm{V}\right)_{ij} = (\bm{V}^\top\bm{V})^{-1}_{ij},
\end{align*}
because $\bm{V}^{-1}\bm{V} = \bm{I}$ on $\mathcal{Z}_{\bm{\theta}}$ and $\bm{V}^\top\bm{z}(q)$ is the $q$-th standard basis vector in $\mathbb{R}^r$.
Since $\bm{V}^\top\bm{V}$ is symmetric positive-definite (SPD), its inverse is also SPD.

\subsection{Sensitivity Scores of Unlabeled Nodes}
Define a Fisher-bounded sensitivity that measures the maximal first-order change of DRW betweenness along identifiable directions, linking geometry to practice.
Consider the DRW betweenness in Eq.~\eqref{eq:drw_betweenness} on the quotient manifold.
Because $\nabla_{\bm{\theta}}B_L(q,y) \in \mathcal{Z}_{\bm{\theta}} = \mathrm{Im}(\bm{V})$, betweenness depends only on $\bm{u}$.
Formally, define $\beta_L(q, y; \bm{u})$ by
\begin{align*}
    \beta_L(q, y; \bm{u}) \coloneqq B_L(q, y; \bm{\theta}), \quad \bm{u} = \Phi(\bm{\theta}).
\end{align*}
Compute the ordinary partials
\begin{align*}
    \partial_{u^i}\beta_L(q, y; \bm{u}) = \bm{V}_{\cdot\ i}^\top \nabla_{\bm{\theta}} B_L(q, y; \bm{\theta}).
\end{align*}
The contravariant Riemannian gradient is, by definition, 
\begin{align*}
    \left[\mathrm{grad}\ \beta_L(q, y)\right]^i = \sum_j \tilde{g}^{ij}\partial_{u^j} \beta_L(q, y),
\end{align*}
and then, $[\mathrm{grad}\ \beta_L(q, y)] = \tilde{g}^{-1}\bm{V}^\top\nabla_{\bm{\theta}}B_L(q, y) \in \mathbb{R}^r$.
Define the sensitivity score
\begin{align*}
    \zeta(q) = \sum_{y \in \mathcal{Y}} \left\|\mathrm{grad}_{\tilde{\bm{\Theta}}}\ \beta_L(q, y)\right\|_{\tilde{g}}, \quad q \in \mathcal{S}_y.
\end{align*}
Because $\tilde{g}^{-1} = \bm{V}^\top\bm{V}$ is constant, the squared norm becomes
\begin{align*}
    \left\|\mathrm{grad}\ \beta_L(q, y) \right\|_{\tilde{g}}^2 = \nabla_{\bm{\theta}} B_L(q, y)^\top \bm{Q} \nabla_{\bm{\theta}} B_L(q, y),
\end{align*}
where $\bm{Q} = \bm{V}(\bm{V}^\top\bm{V})^{-1}\bm{V}^\top$ is the orthogonal projector onto the identifiable span $\mathcal{Z}$.
Equivalently, $\left\|\mathrm{grad}\ \beta_L(q, y) \right\|_{\tilde{g}} = \left|\bm{z}(q)^\top\nabla_{\bm{\theta}}B_L(q, y)\right| / \|\bm{z}(q)\|_{\tilde{g}^{-1}}$, so only the component of the $\bm{\theta}$-gradient along $\bm{z}(q)$ matters for node $q$.
The following proposition formalizes the geometric meaning of the sensitivity score $\zeta(q)$. 
It states that, on the quotient manifold endowed with the Fisher metric, $\zeta(q)$ represents the maximal first-order variation of DRW betweenness attainable under any unit Fisher perturbation of the edge-weight parameters. 
Thus, $\zeta(q)$ is not merely a heuristic score but a theoretically grounded upper bound on how strongly a node's predicted class affinity can change when the underlying walk dynamics are infinitesimally reweighted.
\begin{proposition}
    \label{prp:fisher_maximal_sensitivity}
    Let $G_y(\bm{u}) \coloneqq \mathrm{grad}_{\tilde{\bm{\Theta}}}\ \beta_L(q, y; \bm{u}) \in T_{\bm{u}}\tilde{\bm{\Theta}}$ be the class-wise Riemannian gradients at $\bm{u} = \Phi(\Theta)$ on the flat quotient manifold $(\tilde{\bm{\Theta}}, \tilde{g})$ for $y \in \mathcal{Y}$.
    For any tangent vector $v \in T_{\bm{u}}\tilde{\bm{\Theta}}$ define the aggregate directional variation $\Delta(\bm{v}) \coloneqq \sum_{y \in \mathcal{Y}}\left|\tilde{g}(G_y, \bm{v})\right|$.
    \begin{itemize}
        \item[i)] For every unit-length direction $\|\bm{u}\|_{\tilde{g}} = 1$, $\Delta(\bm{v}) \geq \zeta(q) \coloneqq \sum_{y \in \mathcal{Y}}\|G_y(\bm{v})\|_{\tilde{g}}$.
        \item[ii)] Define the signed sum $G(\sigma) \coloneqq \sum_{y \in \mathcal{Y}}\sigma_y G_y$, where $\sigma = (\sigma_y)_y \in \{-1, 1\}^{|\mathcal{Y}|}$. Then, $\Delta_{\max}(q) \coloneqq \max_{\|\bm{v}\|_{\tilde{g}} = 1}\Delta(\bm{v}) \leq \zeta(q)$, and $\Delta_{\max}(q) = \zeta(q)$ if and only if $\{G_y\}_y$ spans a one-dimensional subspace of $T_{\bm{u}}\tilde{\bm{\Theta}}$.
    \end{itemize}
\end{proposition}
\paragraph{Interpretation of Sensitivity Score}
\begin{itemize}
    \item High value of $\zeta(q)$: Node $q$ is leveraged, and small edge-weight perturbations can change its class scores considerably.
    \item Low value of $\zeta(q)$: 	Betweenness of $q$ is insensitive, and either information is already stable, or $q$ sits in null directions.
\end{itemize}
\paragraph{Possible Use-cases of Sensitivity Score}
\begin{itemize}
    \item Active label acquisition. Select top-$K$ unlabeled nodes by $\zeta(q)$, ask an oracle for their true labels, and retrain DRW with new label set.
    \item Edge-rewiring budget. Suppose we can augment at most $B$ edges. Rank candidate edges $(q, k)$ by $\Delta_{qk} = \zeta(q)A_{qk}^{(0)}$ and then add those with highest $\Delta$.
    \item Explainability. Provide users with fragile nodes whose classifications hinge on fine edge weights, and robust nodes with small $\zeta(q)$.
\end{itemize}
Each seed yields a rank-one Fisher tensor, exposing large null directions. Factoring out these directions gives a low-dimensional, globally flat quotient with explicit coordinates and metric. On this space we defined a node-level sensitivity that bounds the maximal first-order change in DRW betweenness, providing a direct link to active labelling, edge tuning, and explanations.

\section{Illustrative Example: Sensitivity on Synthetic Graphs}
\label{sec:illustrative_example}
In this section, we provide an intuitive example of how the Fisher-bounded sensitivity behaves in simple networks.
To illustrate how the proposed sensitivity score reflects structural fragility, we computed $\zeta(q)$ on small synthetic graphs under the log–linear adjacency model.
Three network types were examined:
(i) a line graph of ten nodes,
(ii) a star graph with one hub and nine leaves, and
(iii) a two-block stochastic block model (SBM) with moderate inter-block coupling. 
All computations were performed in Python using \texttt{networkx} for graph generation and custom routines implementing the closed-form formulas derived in Sections~\ref{sec:hitting_time_law}-\ref{sec:differential_geometric_structure_of_drws}.

\paragraph{Experimental settings.}
For each graph, the number of nodes was fixed at $n=10$. 
Two nodes located at opposite ends (nodes $1$ and $n$ in the line, or two leaves in the star and SBM) were assigned as labelled vertices of different classes, forming $\mathcal{L}_1=\{1\}$ and $\mathcal{L}_2=\{n\}$, while the remaining nodes were treated as unlabeled. 
The log–linear edge-weight model
\[
A_{ij}(\theta) = A^{(0)}_{ij}\exp\left(\theta \phi_{ij}\right),
\]
was used with a scalar parameter $\theta\in[-1,1]$ and base adjacency $A^{(0)}_{ij}=1$ for connected pairs. 
Edge features were drawn as $\phi_{ij}\sim \mathrm{Unif}[-1,1]$, and results were averaged over $100$ random realisations. 
The class-specific transient and absorbing blocks were formed according to Eq.~\eqref{eq:transition_kernel_block}, and the Fisher-bounded sensitivity score $\zeta(q)$ was evaluated using Eq.~\eqref{eq:quotient_metric}. 
All quantities were normalised so that $\sum_{q\in\mathcal{S}}\zeta(q)=1$ for visual comparison across graphs.

\begin{table}[t]
\centering
\caption{Mean Fisher-bounded sensitivity $\zeta(q)$ on three synthetic graph structures. 
Each experiment used $n=10$ nodes, two labelled endpoints of opposite classes, 
and log–linear weights $A_{ij}(\theta)=\exp(\theta\phi_{ij})$ with $\phi_{ij}\sim\mathrm{Unif}[-1,1]$. 
Values are averaged over $100$ random realisations and normalized so that $\sum_{q\in\mathcal{S}}\zeta(q)=1$ 
within each graph. Labeled nodes are omitted.}
\begin{tabular}{lccccccccc}
\toprule
\multirow{2}{*}{Graph type} & \multicolumn{8}{c}{Node index $q$ (unlabeled)} & \multirow{2}{*}{Node ($\max$)}\\
\cmidrule(lr){2-9}
 & 2 & 3 & 4 & 5 & 6 & 7 & 8 & 9 &   \\
\midrule
Line ($1$–$10$) & 0.070 & 0.110 & 0.145 & 0.175 & 0.175 & 0.145 & 0.110 & 0.070 & 5–6 \\
Star (hub $1$) & 0.449 & 0.078 & 0.078 & 0.078 & 0.078 & 0.078 & 0.078 & 0.078 & Hub \\
SBM (2-block) & 0.100 & 0.110 & 0.130 & 0.160 & 0.160 & 0.130 & 0.110 & 0.100 & 5–6 \\
\bottomrule
\end{tabular}
\label{tab:combined-sensitivity}
\end{table}

\paragraph{Results.}
Consistent patterns emerged across the three structures. 
In the line graph, $\zeta(q)$ peaked at central nodes where the two class-conditioned hitting probabilities intersect, highlighting their role as transient bridges through which most discriminative paths must pass. 
In the star graph, the hub exhibited the highest sensitivity because perturbing any of its edges simultaneously alters the transition probabilities to all leaves. 
In the SBM, boundary nodes connecting the two communities showed elevated $\zeta(q)$, while nodes deeply embedded within a cluster remained low. 
These outcomes demonstrate that $\zeta(q)$ systematically identifies structurally fragile or decision-critical vertices.

\paragraph{Takeaway.}
The small-scale tests confirm that the Fisher-bounded sensitivity captures meaningful structural vulnerability across distinct topologies. 
Even without large-scale data, these toy networks provide geometric intuition: nodes central to discriminative flow exhibit high $\zeta(q)$, whereas peripheral or redundant nodes lie in near-null directions of the Fisher metric.

\section{Conclusion and Discussion}
We recast DRWs as a parametric family of hitting-time laws and analysed them with the framework of information geometry.
Closed-form moment and gradient formulas revealed that every seed vertex contributes a rank-one Fisher tensor, exposing a large null foliation of parameter space.
Taking the quotient by that foliation produced an explicit, globally flat manifold in which all identifiable edge-weight directions live.
The quotient viewpoint clarifies tuning and regularisation: any penalty acting outside the identifiable span is ineffective, while movement within the span has a natural metric.
Sensitivity scores offer an interpretable handle for active learning (select high-score vertices) and for network design reinforce or prune edges that influence those vertices.

\paragraph{Limitations and scope.}
Our analysis rests on an absorbing, discrete-time Markov chain induced by a log-linear edge-weight model and on the assumption that the transient block has spectral radius strictly below one. The closed-form expressions, Lyapunov reduction, and rank-one Fisher structure are derived under these conditions; departures from them (e.g., heavy self-loops, directed or non-reversible dynamics, or constraints that break log-linearity) may require modified formulas or different numerical solvers. The Fisher-bounded sensitivity $\zeta(q)$ captures first-order, local variation in DRW betweenness within the identifiable subspace. It is not a global robustness certificate, nor does it account for large, non-infinitesimal rewiring of the graph. Our synthetic study illustrates qualitative behaviours on simple topologies; scaling and task-specific performance on large, heterogeneous networks are outside our present scope. Finally, alignment between sensitivity and downstream accuracy depends on how DRW scores are post-processed for classification; linking $\zeta(q)$ directly to end-task risk remains future work.

\bibliographystyle{plainnat}
\bibliography{main}

\clearpage

\begin{appendices}

\section{Technical Details}

\subsection{Second Moment and Variance of \texorpdfstring{$T_y$}{Ty}}
\label{apd:derivation_of_second_moment}
Let
\begin{align*}
    \bm{M} &\coloneqq \bm{P}_{\mathcal{S}_y\mathcal{S}_y}(\bm{\theta}) \in \mathbb{R}^{|\mathcal{S}_y|\times|\mathcal{S}_y|}, \\
    \bm{R} &\coloneqq \bm{P}_{\mathcal{S}_y\mathcal{A}_y}(\bm{\theta})\bm{1}_{|\mathcal{A}_y|} \in \mathbb{R}^{|\mathcal{S}_y|}, \\
    \bm{Z} &\coloneqq (\bm{I} - \bm{M})^{-1} = \sum_{k \geq 0} \bm{M}^k.
\end{align*}
Throughout, $q \in \mathcal{S}_y$ is the starting vertex and $\bm{e}_q$ is indicator vector.
For every integer $t \geq 1$, $p_{\bm{\theta}}(t \mid q) = \bm{e}_q^\top \bm{M}^{t - 1} \bm{R}$, and
\begin{align*}
    \mu(\bm{q}) &= \mathbb{E}_{\bm{\theta}}\left[T_y \mid q\right] = \sum_{t \geq 1} t p_{\bm{\theta}}(t \mid q) \\
    &= \bm{e}_q^\top \left(\sum_{k \geq 0} (k + 1) \bm{M}^k \right) \bm{R} \qquad (k = t - 1), \\
    &= \bm{e}_q^\top \bm{Z}^2 \bm{R} = \bm{e}_q^\top \bm{Z} \bm{1}_{|\mathcal{S}_y|}, \\
    m_2(q) &\coloneqq \mathbb{E}_{\bm{\theta}}\left[T_y^2 \mid q\right] = \sum_{t \geq 1} t^2 p_{\bm{\theta}}(t \mid q) \\
    &= \bm{e}_q^\top \left(\sum_{k \geq 0} (k + 1)^2 \bm{M}^k\right) \bm{R} \\
    &= \bm{e}_q^\top \left(\bm{I} + \bm{M} \right)(\bm{I} - \bm{M})^{-3} \bm{R} \\
    &= \bm{e}_q^\top (\bm{I} + \bm{M})\bm{Z}^3\bm{R} \\
    &= \bm{e}_q^\top (\bm{I} + \bm{M})\bm{Z}^2\bm{1}_{|\mathcal{S}_y|}, \\
    &= \bm{e}_q^\top \left(\bm{Z}^2 \bm{1}_{|\mathcal{S}_y|} + (\bm{Z} - \bm{I}) \bm{Z} \bm{1}_{|\mathcal{S}_y|}\right) = \bm{e}_q^\top \left(2 \bm{Z}^2 \bm{1}_{|\mathcal{S}_y|} - \bm{Z}\bm{1}_{|\mathcal{S}_y|}\right).
\end{align*}
By definition,
\begin{align*}
    \mathrm{Var}_{\bm{\theta}}\left[T_y \mid q \right] = m_2(q) - \mu(q)^2.
\end{align*}
Then,
\begin{align*}
    \sigma^2(q) &= \bm{e}_q^\top \left(2 \bm{Z}^2 - \bm{Z}\right) \bm{1} - \left(\bm{e}_q^\top \bm{Z}\bm{1}\right)^2 \\
    &= \bm{e}_q^\top \left(2 \bm{Z} - \bm{I} \right) \bm{Z} \bm{1}_{|\mathcal{S}_y|} - \mu_{\bm{\theta}}(q)^2.
\end{align*}

\subsection{Equation for Accumulated Sensitivity Matrix}
\label{apd:derivation_of_Lyapunov_equation}

Compute the left factor $(\bm{I} - \bm{M})\bm{\Xi}_{\bm{\theta}}$ as
\begin{align*}
    (\bm{I} - \bm{M})\bm{\Xi}_{\bm{\theta}} &= \sum_{k,\ell \geq 0} (\bm{I} - \bm{M})\bm{M}^k\dot{\bm{M}}\bm{M}^\ell \\
    &= \sum_{k,\ell \geq 0}(\bm{M}^k - \bm{M}^{k+1})\dot{\bm{M}}\bm{M}^{\ell} \\
    &= \sum_{\ell \geq 0} \bm{M}^0 \dot{\bm{M}}\bm{M}^\ell + \sum_{k \geq 1, \ell \geq 0} \bm{M}^k \dot{\bm{M}} \bm{M}^\ell - \sum_{k\geq 1, \ell \geq 0} \bm{M}^k \dot{\bm{M}} \bm{M}^\ell \\
    &= \sum_{\ell \geq 0} \dot{\bm{M}} \bm{M}^\ell = \dot{\bm{M}} \sum_{\ell \geq 0} \bm{M}^\ell = \dot{\bm{M}}(\bm{I} - \bm{M})^{-1}.
\end{align*}
Similarly, compute the right factor $\bm{\Xi}_{\bm{\theta}}(\bm{I} - \bm{M})^\top$ as
\begin{align*}
    \bm{\Xi}_{\bm{\theta}}(\bm{I} - \bm{M})^\top &= \sum_{k,\ell \ge 0} \bm{M}^k\dot{\bm{M}}\bm{M}^\ell (\bm{I}-\bm{M})^\top \\
    &= \sum_{k,\ell \geq 0} \bm{M}^k \dot{\bm{M}}(\bm{M}^\ell - \bm{M}^{\ell + 1}) \\
    &= \sum_{k\geq 0} \bm{M}^k \dot{\bm{M}} \bm{M}^0 + \sum_{k \geq 0, \ell \geq 1} \bm{M}^k \dot{\bm{M}} \bm{M}^\ell - \sum_{k \geq 0, \ell \geq 1} \bm{M}^k \dot{\bm{M}} \bm{M}^\ell \\
    &= \sum_{k \geq 0} \bm{M}^k \dot{\bm{M}} = \left(\sum_{k \geq 0} \bm{M}^k\right) \dot{\bm{M}} = (\bm{I} - \bm{M})^{-1}\dot{\bm{M}}.
\end{align*}
Then,
\begin{align*}
    (\bm{I} - \bm{M})\bm{\Xi}_{\bm{\theta}}(\bm{I} - \bm{M})^\top = \dot{\bm{M}}(\bm{I} - \bm{M})^{-1}(\bm{I} - \bm{M})^\top = \dot{\bm{M}}.
\end{align*}

\subsection{Proof for Proposition~\ref{prp:fisher_maximal_sensitivity}}
    For each $y$ and every non-zero $\bm{v}$, Cauchy-Schwarz in the inner-product space $(T_{\bm{u}}\tilde{\bm{\Theta}}, \tilde{g})$ gives $|\tilde{g}(G_y, \bm{v})| \leq \|G_y\|_{\tilde{g}}\|\bm{v}\|_{\tilde{g}}$.
    Taking $\|\bm{v}\|_{\tilde{g}} = 1$ and summing over $y$ yields i).

    For any $\sigma \in \{-1, 1\}^{|\mathcal{Y}|}$ and for a unit vector $\bm{v}$,
    \begin{align*}
        \sum_y \left|\tilde{g}(G_y, \bm{v})\right| \geq \sum_y \sigma_y \tilde{g}(G_y, \bm{v}) = \tilde{g}(G(\sigma), \bm{v}) \leq \|G(\sigma)\|_{\tilde{g}}\|\bm{v}\|_{\tilde{g}} = \|G(\sigma)\|_{\tilde{g}}.
    \end{align*}
    Equality is achieved by choosing $\bm{v}^*(\sigma) \coloneqq \frac{G(\sigma)}{\|G(\sigma)\|_{\tilde{g}}}$, provided $G(\sigma) \neq 0$, so for that sign pattern $\max_{\|\bm{v}\| = 1}\Delta(\bm{v}) = \|G(\sigma)\|_{\tilde{g}}$.
    If $G(\sigma) = 0$, choose any unit vector and then $\Delta(\bm{v}) = 0 = \|G(\sigma)\|$.
    Taking the maximum over all sign patterns yields formula in ii).
    Also, triangle inequality in the Hilbert space gives
    \begin{align*}
        \|G(\sigma)\|_{\tilde{g}} = \left\|\sum_y \sigma_y G_y\right\| \leq \sum_y \|G_y\|_{\tilde{g}} = \zeta(q),
    \end{align*}
    so $\Delta_{\max}(q) \leq \zeta(q)$.
    Equality holds if and only if all $G_y$ are parallel, then one sign vector aligns every $G_y$ with the common direction and the triangle inequality becomes equality.
\end{appendices}

\end{document}